\documentclass[letterpaper]{article}
\usepackage{spconf,amsmath,amssymb,graphicx,booktabs,multirow}
\usepackage[hidelinks]{hyperref}

\title{MOSAIC: Montage-Agnostic Self-Distilled Graph Learning for Subject-Independent iEEG Seizure Detection}
\name{Arya Jalali\thanks{\small Corresponding author: arya.jalali@epfl.ch \\ This work was supported in part by the Swiss State Secretariat for Education, Research and Innovation (SERI) under Contract SCR0548363, the SwissChips initiative, and in part by the Swiss National Science Foundation (SNSF) Project Funding 200021-219752. The authors declare no conflicts of interest.}, Mahsa Shoaran}
\address{
EPFL \\
IEM and Neuro-X Institutes\\
Geneva, Switzerland}
\begin{document}
\ninept
\maketitle
\begin{abstract}
Intracranial EEG provides high signal-to-noise ratio and spatial resolution for detecting epileptic activity, but patient-specific electrode placement leads to substantial variation in channel count and montage across subjects. This heterogeneity limits the generalization of existing detectors under subject-independent evaluation. We
propose MOSAIC (MOntage-agnostic Self-distilled Adaptive Inter-channel
Connectivity learning), a lightweight graph-learning framework for
subject-independent iEEG seizure detection. MOSAIC pairs a shared
convolutional channel encoder with a patient-agnostic channel-attention module
that infers a weighted inter-channel adjacency matrix from the encoded features,
requiring no fixed montage graph. An exponential moving average
self-distillation objective regularizes each prediction toward a slowly
updated per-patient teacher. Across four public iEEG corpora (SWEC-ETHZ long-
and short-term, Mayo, FNUSA), MOSAIC outperforms state-of-the-art baselines in
F2 and Cohen's $\kappa$ while learning more stable inter-channel
adjacencies, demonstrating that self-distilled connectivity 
learning transfers across unseen implantations.
\end{abstract}
\section{Introduction}
\label{sec:intro}

Epilepsy is one of the world's most common serious neurological
disorders, affecting an estimated 50 million people
worldwide~\cite{who2024epilepsy}. Roughly a third of patients are
drug-resistant and depend on continuous seizure monitoring and, when
eligible, on surgical resection of  epileptogenic zone or neuromodulation \cite{shin2022neuraltree}. Both still
rest on visual review of scalp EEG or, for more precise localization, of
invasive iEEG (SEEG/ECoG)~\cite{flanary2023reliability} -- a slow,
subjective process that does not scale to modern continuous recordings.

Automated detection has progressed from classifiers based on  handcrafted spectral,
temporal, and spatial features, such as patient-specific
SVMs~\cite{shoeb2010application}, tree-based models \cite{shoaran2018energy, shin2022neuraltree} and hyperdimensional-computing
classifiers~\cite{burrello2019hyperdimensional}, to deep neural networks that learn representations directly from neural signals, ranging from scalp-map recurrent
convolutional networks~\cite{thodoroff2016learning} to CNNs that learn
spatial filters directly over a fixed electrode arrangement, such as
EEGNet~\cite{lawhern2018eegnet} and
EEGWaveNet~\cite{thuwajit2021eegwavenet}. Graph neural networks (GNNs)
further model the montage explicitly, passing messages over an
inter-channel adjacency: DCRNN~\cite{tang2022selfsupervised} derives it
from electrode geometry or signal correlation, and
REST~\cite{afzal2024rest} pairs it with a recurrent residual-state
update. In all these families the layout is fixed at training time and
assumed to hold at deployment.

That assumption fails on iEEG. Although iEEG offers superior
signal-to-noise ratio and localization~\cite{parvizi2018promises},
electrode count and placement are chosen individually per patient, so no
montage graph is shared across subjects and fixed-topology detectors
degrade under subject-independent evaluation. Transformer-based
approaches sidestep this by letting self-attention model inter-channel relationships directly from the signal: MVPFormer~\cite{carzaniga2026foundation}
disentangles content, temporal, and spatial attention across arbitrary
channel counts, and Brant~\cite{zhang2023brant} adds a dedicated
spatial-attention encoder. Both, however, pay the quadratic compute and
inference latency of full self-attention. Inferring the adjacency from a
single short window also inherits the non-stationarity of iEEG
itself~\cite{dikanev2005eeg}: consecutive windows from the same patient,
recorded through the same electrodes, can yield significantly different
graphs.

We introduce MOSAIC, a lightweight graph architecture that infers
inter-channel connectivity per window from a single shared attention
layer, and damps this window-to-window instability with an exponential moving average (EMA)
self-distillation objective that regularizes each inferred graph toward a
slowly updated, patient-specific consensus. MOSAIC therefore adapts to an 
arbitrary, unseen montage while keeping the inferred connectivity stable across windows.

\section{Methodology}
\label{sec:method}

\begin{figure*}[!t]
\centering
\includegraphics[width=\textwidth]{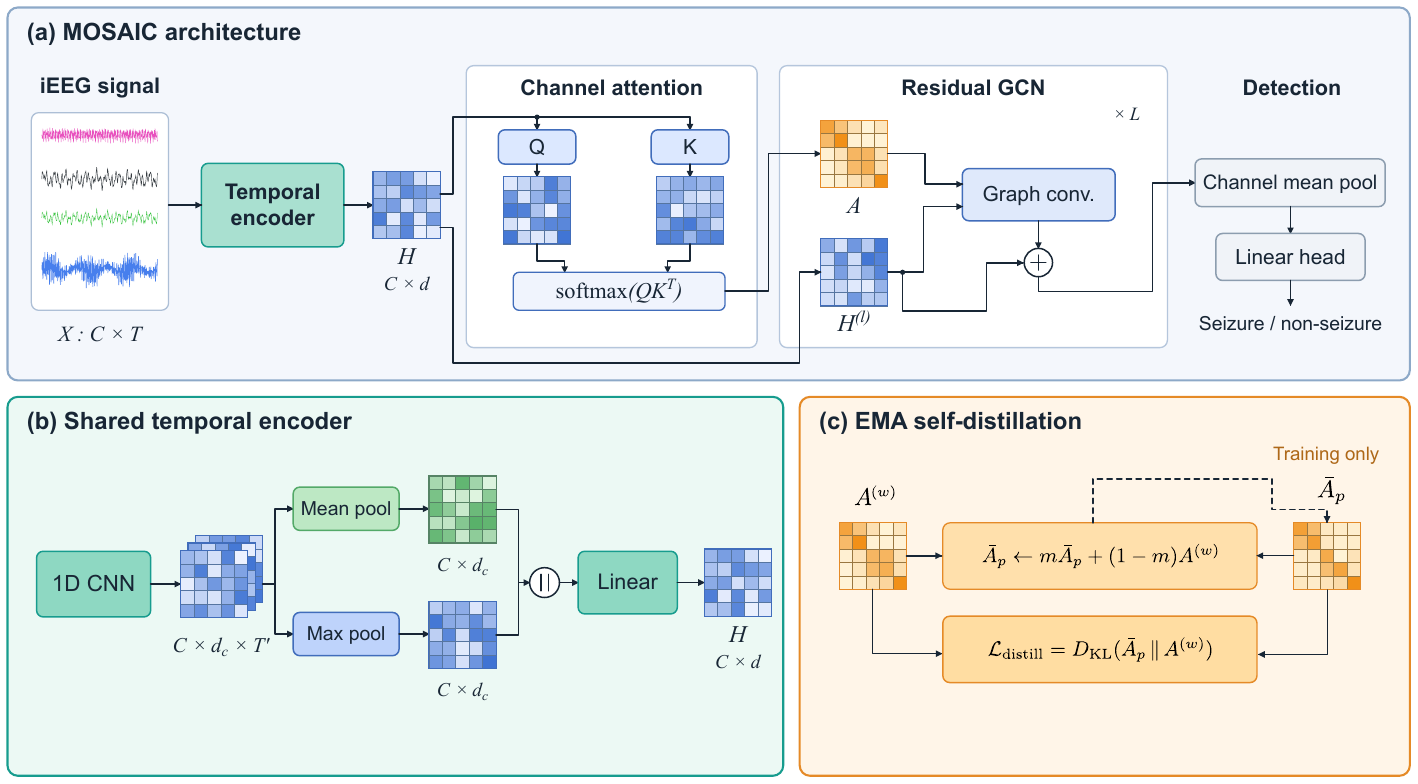}
\caption{Overview of MOSAIC. Each electrode signal in an iEEG window is
encoded independently by a shared temporal encoder into channel-level
node features $H$. Shared query and key projections infer a
row-normalized adjacency $A$. Residual graph-convolution layers propagate
information over the inferred graph, followed by channel pooling
and binary classification. During training, adjacency predictions are
grouped by patient and averaged to update a stop-gradient EMA teacher,
and a KL-divergence term keeps each window-level adjacency consistent
with the patient's historical adjacency structure.}
\label{fig:architecture}
\end{figure*}



\subsection{Problem formulation}
\label{ssec:problem}

Each recording is a continuous multivariate iEEG time series, which we
segment into non-overlapping windows of a fixed duration $t$. Formally, a
window is $x \in \mathbb{R}^{C \times T}$, where $T = t \cdot f_s$ is the
number of timepoints at sampling frequency $f_s$, and $C$ is the number
of channels. Both channel count and electrode placement may vary across patients. Each window carries a binary label $y \in \{0,1\}$, where $y = 1$ marks a
window containing expert-annotated seizure activity. The goal is subject-independent detection: at
test time the model sees a patient $p$ absent from training.

\subsection{Shared temporal encoder}
\label{ssec:encoder}

Every channel's raw window $x_i \in \mathbb{R}^{T}$ is independently
encoded by one shared 1D CNN to extract local temporal features: a stack of $L_{\mathrm{cnn}}$
depthwise-separable convolutional layers, each followed by batch
normalization and ReLU, producing a feature map
$z_i \in \mathbb{R}^{d_c \times T'}$. Inspired by
the channel-pooling strategy in CBAM~\cite{woo2018cbam}, we pool over $T'$ by concatenating mean and max
pooling before projecting back to $d_c$:
\begin{equation}
h_i = W_{\mathrm{pool}}
\Big[\,\textstyle\frac{1}{T'}\sum_{t} z_{i,:,t} \;\Vert\; \max_t z_{i,:,t}\,\Big] \in \mathbb{R}^{d_c};
\label{eq:cnn-pool}
\end{equation}
Max pooling is included specifically to retain sharp, spike-like transients that 
a mean pool alone would average away, which we hypothesize matters for the ictal 
discharges marking seizure onset. Stacking
Eq.~\eqref{eq:cnn-pool} over channels and projecting linearly gives node
features
$H = [h_1; \dots; h_C] W_{\mathrm{in}} \in \mathbb{R}^{C \times d}$
before graph construction.

\subsection{Channel-attention adjacency}
\label{ssec:graph}

We compute a row-stochastic weighted adjacency matrix $A$ as a function of the encoded features $H$ via a
single shared self-attention layer~\cite{vaswani2017attention} with
separate query/key projections $W_Q, W_K$. Following
\cite{henry-etal-2020-query}, queries and keys are $\ell_2$-normalized
before the dot product (QK-norm),
\begin{equation}
\begin{aligned}
\tilde H &= \mathrm{LayerNorm}(H), \\
q_i &= \frac{W_Q^{\top}\tilde{h}_i}{\lVert W_Q^{\top}\tilde{h}_i \rVert_2}, \qquad
k_i = \frac{W_K^{\top}\tilde{h}_i}{\lVert W_K^{\top}\tilde{h}_i \rVert_2},
\end{aligned}
\label{eq:qknorm}
\end{equation}
and a learned scalar temperature $s$ rescales the normalized dot product
before a row-wise softmax,
\begin{equation}
E_{ij} = s\, q_i^{\top} k_j, \qquad
A_{ij} = \frac{\exp(E_{ij})}{\sum_{j'} \exp(E_{ij'})}.
\label{eq:attn-adj}
\end{equation}

\subsection{EMA self-distillation regularizer}
\label{ssec:distill}

The adjacency $A$ of Eq.~\eqref{eq:attn-adj} is recomputed for every
window and is free to fluctuate with local, possibly non-discriminative
signal variation. Since a patient's electrode configuration is fixed, its
channel-relationship structure should remain broadly stable across windows;
we therefore impose a soft structural prior toward this consistency by
maintaining a per-patient exponential moving average
$\bar{A}_p \in \mathbb{R}^{C \times C}$ of the model's own past adjacency
predictions. For a batch, let
$\mathcal{W}_p$ be the set of windows belonging to patient $p$, $n_p =
|\mathcal{W}_p|$, and $A^{(w)}$ the current adjacency for window
$w \in \mathcal{W}_p$. Each window is pulled toward the current teacher
$\bar{A}_p$ via a row-averaged KL-divergence,
\begin{equation}
\ell^{(w)} = \frac{1}{C} \sum_{i=1}^{C} \sum_{j=1}^{C}
\bar{A}_{p,ij} \log \frac{\bar{A}_{p,ij} + \epsilon}{A^{(w)}_{ij} + \epsilon},
\label{eq:distill-window}
\end{equation}
where $\epsilon$ is a small constant for numerical stability. The batch
distillation loss averages Eq.~\eqref{eq:distill-window} within each
patient before averaging over  patients $\mathcal{P}$  in the
batch that already have a teacher ($\bar{A}_p$ is initialized from the first window of a newly seen patient):
\begin{equation}
\mathcal{L}_{\mathrm{distill}} = \frac{1}{|\mathcal{P}|} \sum_{p \in \mathcal{P}}
\frac{1}{n_p} \sum_{w \in \mathcal{W}_p} \ell^{(w)}.
\label{eq:distill}
\end{equation}
After the loss is computed, $\bar{A}_p$ is updated from a detached,
batch-averaged adjacency $\bar{a}_p = \frac{1}{n_p}\sum_{w \in \mathcal{W}_p} \mathrm{sg}[A^{(w)}]$,
\begin{equation}
\bar{A}_p \leftarrow m\,\bar{A}_p + (1-m)\,\bar{a}_p,
\label{eq:ema-update}
\end{equation}
where $m$ is the EMA momentum and $\mathrm{sg}[\cdot]$  denotes the stop-gradient operator.
The total training loss is
\begin{equation}
\mathcal{L} = \mathcal{L}_{\mathrm{CE}} + \lambda\, \mathcal{L}_{\mathrm{distill}},
\label{eq:total-loss}
\end{equation}
with $\lambda = 0$ recovering the unregularized learned-adjacency
ablation.

\subsection{Graph convolution}
\label{ssec:gcn}

Node features $H^{(0)} = H$ are updated by $L$ layers of standard graph
convolution over the learned adjacency $A$, each a linear transform,
ReLU, dropout, and residual connection with layer norm:
\begin{equation}
\begin{aligned}
M^{(\ell)} &= \sigma\big(A\, H^{(\ell-1)} W^{(\ell)}\big), \\
H^{(\ell)} &= \mathrm{LayerNorm}\big(H^{(\ell-1)} + \mathrm{Dropout}(M^{(\ell)})\big).
\end{aligned}
\label{eq:gcn-layer}
\end{equation}
The final layer's node features $H^{(L)}$ are mean-pooled over the
channel dimension, and passed to a linear
classifier producing the window-level logit.

\section{Experiments and Results}
\label{sec:experiments}

\subsection{Experimental Settings}
\label{ssec:setup}

\noindent\textbf{Datasets.} We evaluate MOSAIC on four public iEEG corpora,
following the evaluation procedure of \cite{carzaniga2026foundation}. The
SWEC-ETHZ short-term~\cite{burrello2019hyperdimensional} and
long-term~\cite{burrello2019laelaps} corpora are continuous,
multi-channel recordings with annotated seizure intervals, downsampled to
256\,Hz, and we segment them into non-overlapping $t = 2$\,s windows
labelled according to overlap with those seizure intervals. Mayo and
FNUSA~\cite{nejedly2020multicenter} instead consist of fixed-length,
single-channel clips (3\,s; 193k clips from 25 patients and 155k from 14
patients, respectively, downsampled to 2500\,Hz) annotated as pathological activity, physiological activity, powerline
noise, or artifact; we do not remove any noise from these corpora and merge
powerline noise and artifact into
the physiological class, to simulate a real clinical setup.

\noindent\textbf{Baselines.} We compare MOSAIC against
EEGNet~\cite{lawhern2018eegnet}, DCRNN~\cite{tang2022selfsupervised},
SEEGNet~\cite{wang2022seegnet}, DMNet~\cite{tu2024dmnet}, a lightweight
CNN baseline~\cite{si2023lightweight}, and a Multitask
Transformer~\cite{sun2025multitask}. SEEGNet and DMNet are designed for
single-channel iEEG and are therefore not applicable to the
multi-channel SWEC-ETHZ corpora (marked as -- in Table~\ref{tab:results}).

\noindent\textbf{Metrics.} 
We report
Cohen's $\kappa$ \cite{carzaniga2026foundation} on SWEC-ETHZ and macro-F1, F2, and accuracy on all four corpora. F2 places greater weight on recall than precision, reflecting the higher cost of missed seizures.

\noindent\textbf{Protocol and implementation.} On the SWEC-ETHZ corpora
we adopt a leave-one-patient-out protocol: for each fold, all patients
but two are used for training, one is held out for validation, and the
remaining patient is used for testing. Mayo and FNUSA are highly imbalanced
across individual patients, so we instead  follow the grouped leave-one-group-out protocol of~\cite{wang2022seegnet}. Being single-channel, these two corpora also make the channel-attention adjacency
of Eq.~\eqref{eq:attn-adj} degenerate, so on those two corpora MOSAIC
reduces exactly to its shared encoder ($\lambda = 0$). To ensure a fair
comparison, we tune
the hyperparameters of every baseline on validation loss. All models are
implemented in PyTorch~2.11.0 (CUDA~13.0) and trained on two NVIDIA RTX
PRO 6000 Blackwell GPUs with Adam~\cite{kingma2015adam} at a learning
rate of $10^{-3}$ for 10 epochs, with EMA momentum $m = 0.99$ and
distillation weight $\lambda = 0.1$ on the SWEC-ETHZ corpora. All
reported results are averaged over 5 random seeds.

\subsection{Main Results}
\label{ssec:results}

\begin{table*}[!t]
\centering
\caption{Seizure detection results (\%) across four public iEEG datasets.
F2, m-F1 (macro-F1), and accuracy are reported for all datasets; Cohen's
$\kappa$ is additionally reported for the two SWEC-ETHZ corpora.}
\label{tab:results}
\small
\setlength{\tabcolsep}{2pt}
\begin{tabular*}{\textwidth}{@{\extracolsep{\fill}}lcccccccccccccc}
\toprule
\multirow{2}{*}{Model} & \multicolumn{3}{c}{Mayo} & \multicolumn{3}{c}{FNUSA} & \multicolumn{4}{c}{SWEC (short-term)} & \multicolumn{4}{c}{SWEC (long-term)} \\
\cmidrule(lr){2-4} \cmidrule(lr){5-7} \cmidrule(lr){8-11} \cmidrule(lr){12-15}
 & F2 & m-F1 & acc. & F2 & m-F1 & acc. & $\kappa$ & F2 & m-F1 & acc. & $\kappa$ & F2 & m-F1 & acc. \\
\midrule
SEEGNet              & 62.4          & 74.9          & 85.0          & 80.4          & 85.4          & 88.8          & --            & --            & --            & --            & --            & --            & --            & --            \\
DMNet                & 73.1          & 79.8          & 87.3          & 79.2          & 83.0          & 86.7          & --            & --            & --            & --            & --            & --            & --            & --            \\
EEGNet               & 59.2          & 72.3          & 82.4          & 77.0          & 81.7          & 85.4          & 34.9          & 47.3          & 64.8          & 78.2          & 20.1          & 24.4          & 58.1          & 91.9          \\
DCRNN                & 65.7          & 74.3          & 83.7          & 79.2          & 85.0          & 88.5          & 34.5          & 54.6          & 66.4          & 73.4          & 19.8          & 32.8          & 65.5          & 96.9          \\
Lightweight CNN      & 75.8          & \textbf{83.9} & \textbf{91.5} & 82.0          & 86.2          & 89.4          & 46.6          & 60.3          & 70.2          & 77.4          & 40.2          & 44.3          & 69.4          & 96.4          \\
Multitask Transformer & \textbf{78.0} & \textbf{83.9} & 91.2          & 82.7          & 86.2          & 89.3          & 49.3          & 60.9          & 71.4          & 79.0          & 47.3          & 49.1          & 73.2          & 97.2          \\
\textbf{MOSAIC (proposed)} & 77.2 & 83.7          & 91.2          & \textbf{85.2} & \textbf{88.2} & \textbf{91.2} & \textbf{51.1} & \textbf{64.1} & \textbf{73.1} & \textbf{79.7} & \textbf{50.6} & \textbf{52.1} & \textbf{75.2} & \textbf{98.6} \\
\bottomrule
\end{tabular*}
\end{table*}

Table~\ref{tab:results} summarizes seizure-detection performance across the four
datasets. MOSAIC outperforms every baseline on both SWEC-ETHZ corpora,
improving $\kappa$ over the strongest baseline by 1.8 points on the
short-term corpus and 3.3 points on the long-term one, while training 3$\times$ 
faster than the Multitask Transformer (100 vs.\ 300 min). This is because MOSAIC infers 
the adjacency once and reuses it across all graph layers instead of recomputing 
attention at every layer. The margin is far larger still against the graph-based
baseline DCRNN~\cite{tang2022selfsupervised}: +16.6~$\kappa$ on the short-term corpus
and +30.8 on the long-term one. On the
single-channel Mayo and FNUSA, where MOSAIC reduces to its shared
encoder (Section~\ref{ssec:setup}), it dominates on FNUSA and is
competitive on Mayo, within 0.8 points of the best F2 and macro-F1. The learned
adjacency and its regularizer are active only on the multi-channel SWEC-ETHZ
corpora, precisely the variable-montage, subject-independent setting that
clinical deployment demands, and it is there that the F2 margins are largest.

\subsection{Ablation Experiments}
\label{ssec:ablation}

\begin{figure*}[!t]
\centering
\includegraphics[width=0.74\textwidth]{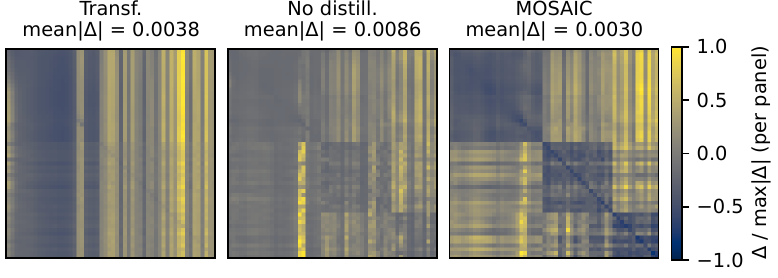}
\caption{Channel-attention deviations from uniform for one held-out
patient. We plot $\Delta = A - \frac{1}{C}\mathbf{1}\mathbf{1}^{\top}$,
averaged over 60 windows. It measures departure  from uniform
attention (rows: query; columns: key). Panels use their own maximum
absolute deviation for color scaling; mean\,$|\Delta|$ is shown above each.
The Transformer concentrates on fixed key columns; the undistilled model
has the largest, least regular deviations; MOSAIC has smaller, coherent
channel blocks. All three suppress the diagonal, already supplied by
the residual in Eq.~\eqref{eq:gcn-layer}.}
\label{fig:heatmap}
\end{figure*}

\begin{table}[!t]
\centering
\caption{Ablation study (\%) isolating the temporal encoder, the EMA
self-distillation regularizer, and the learned adjacency itself, on the
two SWEC-ETHZ corpora.}
\label{tab:ablation}
\small
\begin{tabular*}{\linewidth}{@{\extracolsep{\fill}}lcccc@{}}
\toprule
\multicolumn{5}{c}{SWEC-ETHZ short-term} \\
\midrule
Variant & $\kappa$ & F2 & m-F1 & acc. \\
\midrule
\textbf{MOSAIC} & \textbf{51.1} & \textbf{64.1} & \textbf{73.1} & 79.7 \\
w/o temporal encoder & 31.2 & 47.8 & 62.4 & 72.4 \\
w/o self-distillation & 48.3 & 62.4 & 72.0 & \textbf{79.8} \\
Uniform adjacency & 48.8 & 62.4 & 71.3 & 77.7 \\
\midrule
\multicolumn{5}{c}{SWEC-ETHZ long-term} \\
\midrule
Variant & $\kappa$ & F2 & m-F1 & acc. \\
\midrule
\textbf{MOSAIC} & \textbf{50.6} & \textbf{52.1} & \textbf{75.2} & \textbf{98.6} \\
w/o temporal encoder & 17.9 & 17.4 & 58.3 & 96.1 \\
w/o self-distillation & 46.4 & 47.0 & 72.9 & 97.7 \\
Uniform adjacency & 46.9 & 46.8 & 73.0 & 98.4 \\
\bottomrule
\end{tabular*}
\end{table}

\begin{figure}[!htb]
\centering
\includegraphics[width=0.6\linewidth]{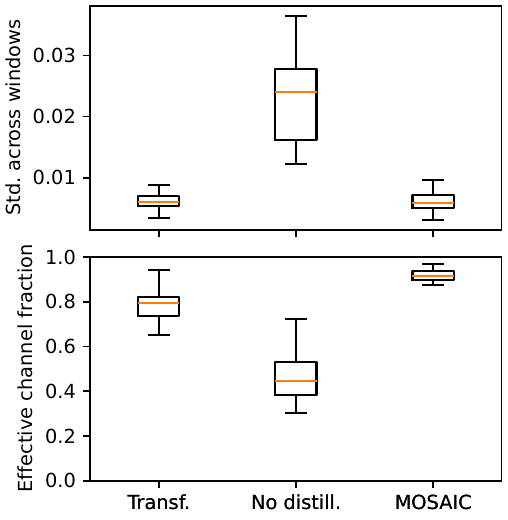}
\caption{Self-distillation keeps the channel-attention adjacency stable
and diffuse ($n=16$ patients). Top: standard deviation of attention
weights across windows within a patient, which removing self-distillation
raises roughly fourfold ($0.006 \rightarrow 0.024$). Bottom: effective
channel fraction $\exp(H_i)/C$, $H_i = -\sum_j A_{ij}\log A_{ij}$.
Undistilled attention concentrates on fewer than half of the channels; MOSAIC holds
near $0.92$.}
\label{fig:consistency}
\end{figure}

Table~\ref{tab:ablation} evaluates three components of MOSAIC:
 the shared temporal encoder (Section~\ref{ssec:encoder}),
 the EMA self-distillation regularizer
(Section~\ref{ssec:distill}, $\lambda = 0$), and  the learned
adjacency. 
Removing the temporal encoder causes the largest collapse: without an informative per-channel representation, the attention module has no signal from which to infer a meaningful adjacency.
The remaining two rows isolate the contribution of the learned adjacency and its regularization. An
unregularized learned adjacency does not improve over the uniform adjacency
($\kappa$ 48.3 vs.\ 48.8, identical F2), so learning the adjacency provides no benefit on its own; only the regularized adjacency improves over the uniform control,
by 2.3~$\kappa$ -- exceeding MOSAIC's own margin over the strongest
baseline in Table~\ref{tab:results}, which the uniform-adjacency control would not
surpass. The improvement is therefore attributable to the stabilization
rather than to the attention mechanism itself. Figure~\ref{fig:heatmap} shows
that the regularized adjacency remains near-uniform, departing from it in 
coherent query–key blocks, whereas the Transformer's deviation is dominated 
by column structure with near-identical rows: every query channel attends 
to essentially the same keys. Figure~\ref{fig:consistency} shows it is also
far more stable across windows than the undistilled adjacency, and more
diffuse than both.

\section{Conclusion}
\label{sec:conclusion}

We presented MOSAIC, a graph neural network for subject-inde\-pendent
iEEG seizure detection that replaces fixed correlation graphs and per-patient adjacency parameters with a shared channel-attention layer that predicts a window-specific adjacency directly from encoded features, allowing adjacency construction itself to generalize to unseen patients. An EMA
self-distillation regularizer stabilizes these window-level adjacencies across windows by
pulling each prediction toward a slowly updated, per-patient average of
the model's own past outputs, without any external supervision. Across
four public iEEG corpora, MOSAIC matches or exceeds competitive
baselines, its undistilled ablation, and a uniform-adjacency control. Because the EMA teacher is only ever a
smoothed copy of the model's own predictions, however, the distillation
loss acts as a pure variance reducer: it reduces variation in the adjacency but cannot correct a systematic bias, such as persistent concentration on a single dominant channel, the way an external supervisory signal could. Future work
therefore includes extending the distillation target beyond a per-patient
average -- e.g., incorporating external supervision -- and evaluating on
scalp EEG.
\par

\raggedbottom
\par\medskip
{\centering\bfseries COMPLIANCE WITH ETHICAL STANDARDS\par}
\nopagebreak
\smallskip
\noindent
This research was conducted retrospectively on de-identified human iEEG
data released in open access by the original providers
\cite{burrello2019hyperdimensional, burrello2019laelaps,
nejedly2020multicenter}. The Mayo and FNUSA recordings were collected
under written informed consent with approval from the Mayo Clinic
Institutional Review Board, the St.\ Anne's University Hospital Research
Ethics Committee, and the Ethics Committee of Masaryk University; the
SWEC-ETHZ recordings are distributed in anonymized form by the
Sleep-Wake-Epilepsy-Center, Inselspital Bern. No new human data were
collected, and no additional ethical approval was required.


\end{document}